\documentclass[11pt,a4paper]{article}
\usepackage{times,latexsym}
\usepackage{url}

\usepackage{latexsym}
\usepackage{epstopdf}
\usepackage{longtable}
\usepackage{tabularx}
\usepackage{enumerate}
\usepackage{mathtools}
\usepackage{tikz}
\usepackage{float}
\usepackage{fancyvrb}
\usepackage{pdflscape}
\usepackage{multirow}
\usepackage{enumitem}
\usepackage{booktabs}
\usepackage{algorithm}
\usepackage{algpseudocode}
\usepackage{xcolor}
\usepackage{xr}
\usepackage{url}
\usepackage{CJKutf8}

\newcommand{\ignore}[1]{}
\usepackage[T1]{fontenc}
\usepackage{xspace,mfirstuc,tabulary}

\usepackage[acceptedWithA]{tacl2018v2}

\title{Better Document-Level Machine Translation with Bayes' Rule}

\author{
 Lei Yu$^1$, Laurent Sartran$^1$,  Wojciech Stokowiec$^1$, \\
 \bf{Wang Ling$^1$, Lingpeng Kong$^1$, Phil Blunsom$^{1,2}$, Chris Dyer$^1$} \\
 $^1$DeepMind, $^2$University of Oxford \\
 {\sf \{leiyu, lsartran, wstokowiec, lingwang, lingpenk, pblunsom, cdyer\}@google.com} \\
}

\date{}

\begin{document}
\maketitle
\begin{abstract}
  We show that Bayes' rule provides an effective mechanism for creating document translation models that can be learned from only parallel sentences and monolingual documents---a compelling benefit as parallel documents are not always available. In our formulation, the posterior probability of a candidate translation is the product of the unconditional (prior) probability of the candidate output document and the ``reverse translation  probability'' of translating the candidate output back into the source language. Our proposed model uses a powerful autoregressive language model as the prior on target language documents, but it assumes that each sentence is translated independently from the target to the source language. Crucially, at test time, when a source document is observed, the document language model prior induces dependencies between the translations of the source sentences in the posterior. The model's independence assumption not only enables efficient use of available data, but it additionally admits a practical left-to-right beam-search algorithm for carrying out inference. Experiments show that our model benefits from using cross-sentence context in the language model, and it outperforms existing document translation approaches.
\end{abstract}

\section{Introduction}
There have been many recent demonstrations that neural language models based on transformers~\citep{VaswaniSPUJGKP17,DaiYYCLS19} are capable of learning to generate remarkably coherent documents with few~\citep{zellers2019neuralfakenews} or no~\citep{radford2019language} conditioning variables. Despite this apparent generation ability, in practical applications, unconditional language models are most often used to provide representations for natural language understanding applications~\citep{DBLP:conf/naacl/DevlinCLT19,DBLP:journals/corr/abs-1906-08237,Peters:2018}, and how to use them for conditional generation applications remains an open question.

Our hypothesis in this work is that Bayes' rule provides an effective way to leverage powerful unconditional document language models to improve a conditional task: machine translation. The application of Bayes' rule to transform the translation modeling problem $p(\boldsymbol{y} \mid \boldsymbol{x})$, where $\boldsymbol{y}$ is the target language, and $\boldsymbol{x}$ is the source language, has a long tradition and was the dominant paradigm in speech and language processing for many years~\citep{DBLP:journals/coling/BrownPPM94}, where it is often called a ``noisy channel'' decomposition, by analogy to an information theoretic conception of Bayes' rule.

While several recent papers have demonstrated that the noisy channel decomposition has benefits when translating sentences one-by-one~\citep{yu:2017,yee-etal-2019-simple,DBLP:conf/wmt/NgYBOAE19}, in this paper we show that this decomposition is particularly suited to tackling the problem of translating complete documents. Although using cross-sentence context and maintaining cross-document consistency has long been recognized as essential to the translation problem~\citep[\textit{inter alia}]{TiedemannS17,BawdenSBH18}, operationalizing this in models has been challenging for several reasons. Most prosaically, parallel documents are not generally available (while parallel sentences are much more numerous), making direct estimation of document translation probabilities challenging. More subtly, documents are considerably more diverse than sentences, and models must be carefully biased so as not to pick up spurious correlations.

Our Bayes' rule decomposition~(\S\ref{sec:model}) permits several innovations that enable us to solve these problems. Rather than directly modeling the conditional distribution, we rewrite it as $p(\boldsymbol{y} \mid \boldsymbol{x}) \propto p(\boldsymbol{y}) \times p(\boldsymbol{x} \mid \boldsymbol{y})$. This changes the learning problem from estimating a single complex conditional distribution to learning two different distributions: a language model $p(\boldsymbol{y})$, which provides unconditional estimates of the output (in this paper, documents); and $p(\boldsymbol{x} \mid \boldsymbol{y})$, which provides the probability of translating a candidate output $\boldsymbol{y}$ into the (observed) source document $\boldsymbol{x}$.

As we will discuss below, although the problems of estimating $p(\boldsymbol{y} \mid \boldsymbol{x})$ and $p(\boldsymbol{x} \mid \boldsymbol{y})$ are formally similar, independence assumptions made in $p(\boldsymbol{x} \mid \boldsymbol{y})$ are less statistically costly than they might otherwise be since, at test time, we will be conditioning on $\boldsymbol{x}$ and reasoning about a posterior distribution over $\boldsymbol{y}$, which will be jointly dependent on all (conditionally independent) parts of $\boldsymbol{x}$.
This statistical fact---which is the same trick that gives na{\"\i}ve Bayes classifiers their expressiveness and ease of estimation---permits us to assume independence between sentence translations in the reverse translation model, and therefore to use parallel sentences (rather than parallel documents) to train it. In the posterior, we thus have an implicit estimate of a document-level translation system, even though we made no use of parallel documents when estimating the prior or likelihood models. This is particularly useful since parallel sentences are much more readily available than parallel documents. A second benefit of our approach is that the unconditional language model can be estimated from nonparallel data, which exists in vast quantities. 
\ignore{
The benefits of strong language models estimated from nonparallel data has long been recognized in translation~\citep{brants-etal-2007-large}, and more recently in noisy channel approaches to sentence-level translation based on neural network component distributions~\citep{yu:2017,yee-etal-2019-simple,DBLP:conf/wmt/NgYBOAE19}.
}

Although the noisy channel model is ideal for exploiting the data resources that naturally exist in the world (large corpora of parallel but independent sentences, and large corpora of monolingual documents), we are faced with a much harder decoding problem~(\S\ref{sec:decoding}). To address this problem, we propose a new beam-search algorithm, exploiting the fact that our document language model operates left-to-right, and our reverse translation model treats sentences independently. The search is guided by a proposal distribution that provides candidate continuations of a document prefix, and these are reranked according to the posterior distribution. In particular, we compare two proposal models: one based on estimates of independent sentence translations~\citep{VaswaniSPUJGKP17} and one that conditions on the source document context~\citep{ZhangLSZXZL18}. While closely related, our algorithm is much simpler and faster than that proposed in \citet{yu:2017}. Rather than using a specially designed channel model \citep{yu:2016} which is limited in processing long sequences like documents, our conditional sentence independence assumptions allow us to use any sequence-to-sequence model as the channel model, making it a better option for document-level translation.

To explore the performance of our proposed model, we focus on Chinese--English translation, following a series of papers on document translation~\citep{ZhangLSZXZL18,WerlenRPH18,TuLSZ18,DBLP:conf/aaai/XiongH0W19}. Although in general it is unreasonable to expect that independent translations of sentences would lead to coherent translations of documents, the task of translating Chinese into English poses some particularly acute challenges.
As Chinese makes fewer inflectional distinctions than English does, and the relevant clues for predicting, e.g., what tense an English verb should be in, or whether an English noun should have singular or plural morphology, may be spread throughout a document, it is crucial that extra-sentential context is used.

Our experiments (\S\ref{sec:experiments}) explore: (1) different approaches to reranking, (2) different independence assumptions when modeling documents (i.e., whether sentences are generated independently or not), (3) different amounts of language modeling data, and (4) different proposal models. Briefly summarized, we find that document-context language models significantly improve the translation quality obtained with our system, both in terms of BLEU scores, and in terms of a human evaluation. Targeted error analysis demonstrates the document prior is capable of enforcing consistency of tense and number and lexical choice across documents.

\section{Model Description}
\label{sec:model}
We define $\underline{\boldsymbol{x}} = (\boldsymbol{x}^1, \boldsymbol{x}^2, \ldots, \boldsymbol{x}^I)$ as the source document with $I$ sentences, and similarly, $\underline{\boldsymbol{y}} = (\boldsymbol{y}^1, \boldsymbol{y}^2, \ldots, \boldsymbol{y}^J)$ as the target document with $J$ sentences. During the (human) translation process, translators may split or recombine sentences, but we will assume that $I=J$.\footnote{Size mismatches are addressed by merging sentences using sentence alignment algorithms~\cite{gale:1993}.} Let $\boldsymbol{x}^i = (x_1^i, x_2^i, \ldots, x_M^i)$ represent the $i$th sentence in the document, consisting of $M$ words; likewise $\boldsymbol{y}^i = (y_1^i, y_2^i, \ldots, y_N^i)$ denote the $i$th sentence in the target document, containing $N$ words.

The translation of a document $\underline{\boldsymbol{x}}$ is determined by finding the document $\hat{\underline{\boldsymbol{y}}}$, where $p(\hat{\underline{\boldsymbol{y}}} \mid \underline{\boldsymbol{x}})$ is optimal.

\begin{equation}
    \hat{\underline{\boldsymbol{y}}} = \arg \max_{\underline{\boldsymbol{y}}} p(\underline{\boldsymbol{y}} \mid \underline{\boldsymbol{x}}).
\end{equation}

\begin{figure}
	\centering
	\includegraphics[scale=0.6]{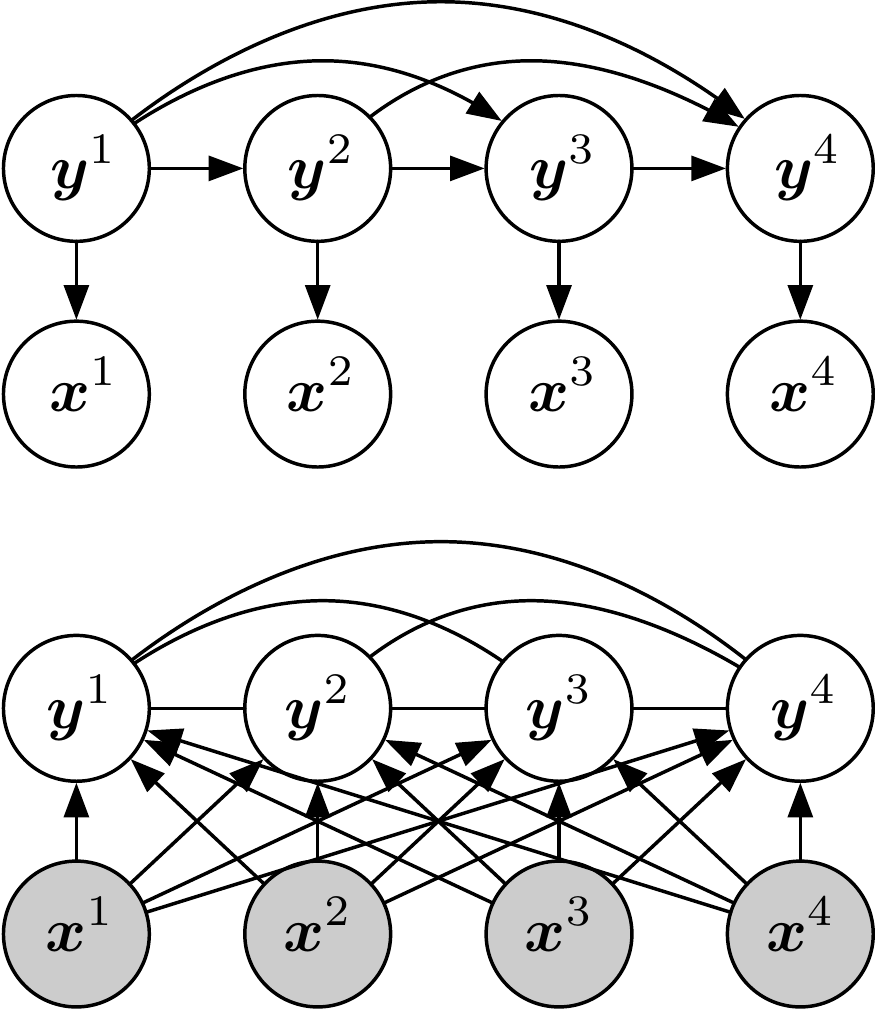}
	\caption{Graphical model showing the factorization of our noisy channel model where $\boldsymbol{y}^i$ indicates the $i$th target language sentence and $\boldsymbol{x}^i$ indicates the $i$th source language sentence. In the prior (top) the target sentences (the $\boldsymbol{y}^i$'s) only influence the corresponding source sentence and therefore can be learned and modeled independently, but at test time (bottom), when the target is not observed, each $\boldsymbol{y}^i$ depends on every $\boldsymbol{x}^i$.}
	\label{fig:graphical_model}
\end{figure}

Instead of modeling the probability $p(\underline{\boldsymbol{y}} \mid \underline{\boldsymbol{x}})$ directly, we factorize it using Bayes' rule:
\begin{equation}
    \begin{split}
        \hat{\underline{\boldsymbol{y}}} &= \arg \max_{\underline{\boldsymbol{y}}} \frac{p(\underline{\boldsymbol{x}} \mid \underline{\boldsymbol{y}}) \times p(\underline{\boldsymbol{y}})}{p(\underline{\boldsymbol{x}})} \\
        &= \arg \max_{\underline{\boldsymbol{y}}} \underbrace{p(\underline{\boldsymbol{x}} \mid \underline{\boldsymbol{y}})}_{\textrm{channel model}} \times \underbrace{p(\underline{\boldsymbol{y}})}_{\textrm{language model}}.
    \end{split}
    \label{bayes}
\end{equation}
We further assume that sentences are independently translated, and that the sentences are generated by a left-to-right factorization according to the chain rule. Therefore, we have
\begin{equation}
    \hat{\underline{\boldsymbol{y}}} \approx \arg \max_{\underline{\boldsymbol{y}}} \prod_{i=1}^{|\underline{\boldsymbol{x}}|} p(\boldsymbol{x}^i \mid \boldsymbol{y}^i) \times p(\boldsymbol{y}^i \mid \underline{\boldsymbol{y}}^{<i}),
    \label{iid_bayes}
\end{equation}
where $\underline{\boldsymbol{y}}^{<i} = (\boldsymbol{y}^1, \ldots, \boldsymbol{y}^{i-1})$ denotes a document prefix consisting of the first $i-1$ target sentences. Thus conceived, this is a generative model of parallel documents that makes a particular independence assumption; we illustrate the corresponding graphical model on the top of Figure~\ref{fig:graphical_model}.

\subsection{Impact of the Conditional Independence Assumption}
At first glance, the conditional independence assumption we have made might seem to be the very independence assumption that bedevils conventional sentence-based approaches to document translation---translations of sentence $i$ appear to be uninfluenced by the translation of any sentence $j \ne i$. However, while this is the case during training, this is \emph{not} the case at test time. Then, we will be conditioning on the $\boldsymbol{x}_i$'s (the source language sentences), and reasoning about the posterior distribution over the ``underlying'' $\boldsymbol{y}_i$'s. By conditioning on the child variables, conditional dependencies between all $\boldsymbol{y}_i$'s and between each $\boldsymbol{y}_i$ and all $\boldsymbol{x}_i$'s are created~\citep{shachter:1998}. The (in)dependencies that are present in the posterior distribution are shown in the bottom of Figure~\ref{fig:graphical_model}.

Thus, although modeling $p(\underline{\boldsymbol{y}} \mid \underline{\boldsymbol{x}})$ or $p(\underline{\boldsymbol{x}} \mid \underline{\boldsymbol{y}})$ would appear to be superficially similar, the statistical impact of making a conditional independence assumption is quite different. This is fortunate, as it makes it straightforward to use parallel sentences, rather than assuming we have parallel documents which are less often available~\citep[\textit{inter alia}]{voita-etal-2019-good,ZhangLSZXZL18,MarufMH19}. Finally, since we only need to learn to model the likelihood of sentence translations (rather than document translations), the challenges of guiding the learners to make robust generalizations in direct document translation models~\citep[\textit{inter alia}]{voita-etal-2019-good,ZhangLSZXZL18,MarufMH19} are neatly avoided.

\subsection{Learning}

We can parameterize the channel probability $p(\boldsymbol{x}^i \mid \boldsymbol{y}^i)$ using any sequence-to-sequence model and parameterize the language model $p(\boldsymbol{y}^i \mid \underline{\boldsymbol{y}}^{<i})$ using any language model.
It is straightforward to learn our model: we simply optimize the channel model and the language model separately on parallel data and monolingual data, respectively.
We remark that it is a significant practical advantage of this parameterization that we can retrain the channel and language models independently, for example if we acquire more monolingual data, or use different language models with the same channel model conditioned on the domain of the source text.

\section{Decoding}
\label{sec:decoding}

Because of the global dependencies in the posterior distribution, decoding in our model is computationally complex. On one hand, similar to the decoding problem faced in standard sequence-to-sequence models, we must search over the space of all possible outputs with a model that makes no Markov assumptions. On the other hand, unlike traditional models, we have to have a complete $\boldsymbol{y}_i$ before we can compute $p(\boldsymbol{x}_i \mid \boldsymbol{y}_i)$, making greedy and near-greedy algorithms ineffective. To address this issue, we use an auxiliary proposal model $q(\underline{\boldsymbol{y}} \mid \underline{\boldsymbol{x}})$, that approximates the posterior distribution using a direct model, to focus our search on promising parts of the output space.

\begin{figure*}
	\centering
	\includegraphics[scale=0.58]{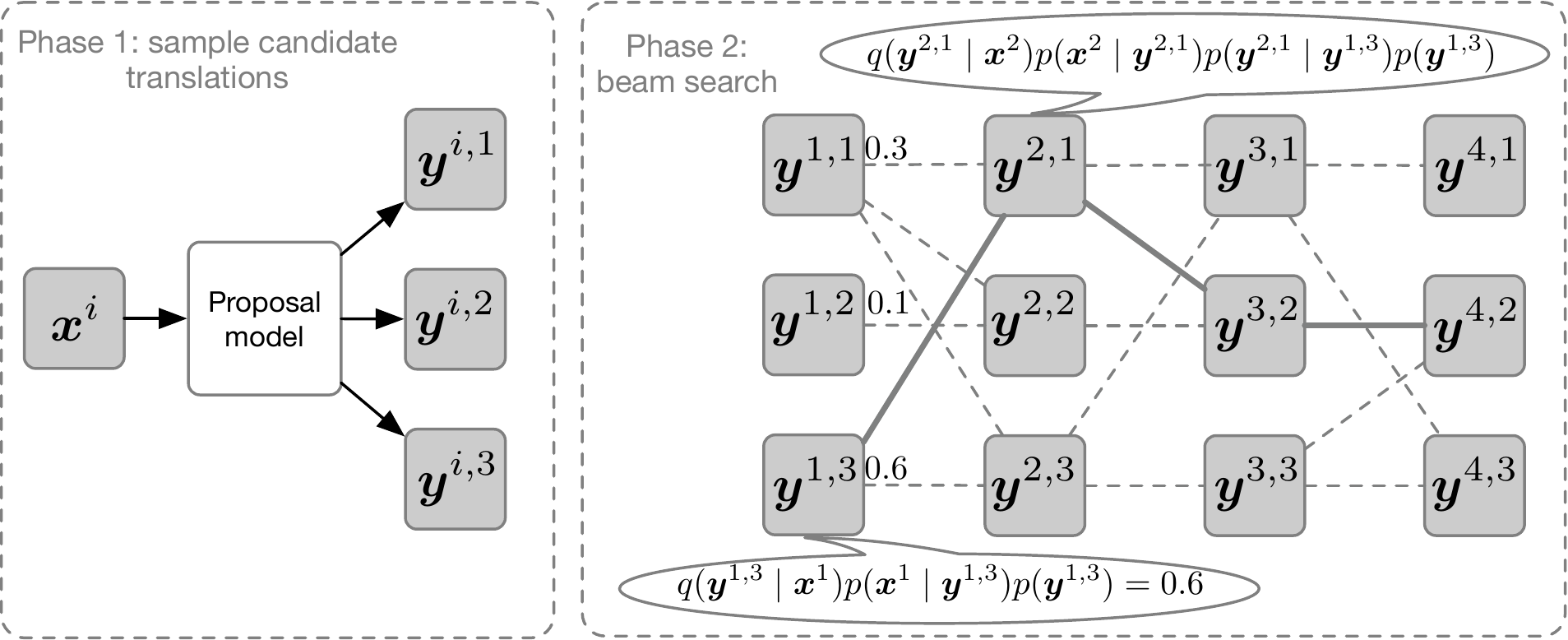}
	\caption{The decoding process. In Phase 1, the auxiliary proposal model generates candidate translations (3 candidates in the diagram) for each sentence in the document (containing 4 sentences). In Phase 2, beam search is employed to search for the best path from the candidate translations.
	}
	\label{doc_mt}
\end{figure*}

Because of the autoregressive factorization of the language model ($p_{\textrm{LM}}$), and the independent sentence translation assumption in the channel model ($p_{\textrm{TM}}$), we can carry out the reranking process using a left-to-right beam search strategy with the aid of our proposal model ($q$). Figure \ref{doc_mt} illustrates the decoding process. For an input document of $\ell$ sentences, we let the proposal model propose $K$ candidate translations for each sentence in the document.\footnote{Our proposal model can optionally use document context on the source (conditioning) side, but sentences are generated independently.} We then search for the best document path through this lattice---or confusion network~\citep{mangu:2000}---of candidate sentence translations. To do so, we maintain a beam of the $B$ active hypotheses (i.e., when considering the $i$th sentence, the prefix consists of $i-1$ sentences), and we consider the proposal's $K$ one-sentence extensions (which we write $\boldsymbol{y}^i$). We retain $B$ partial translations from the $K \times B$ candidates according to the following linear objective,
\begin{align}        \mathcal{O}(\underline{\boldsymbol{x}}, \underline{\boldsymbol{y}}^{<i}, \boldsymbol{y}^i) = &\lambda_1 \log q(\boldsymbol{y}^{i} \mid \underline{\boldsymbol{x}}) + \nonumber \\
& \log p_{\textrm{LM}}(\boldsymbol{y}^{i} \mid \underline{\boldsymbol{y}}^{< i}) + \nonumber \\
 & \lambda_2 \log p_{\textrm{TM}}(\boldsymbol{x}^{i} \mid \boldsymbol{y}^{i}) + \lambda_3 |\boldsymbol{y}^{i}| + \nonumber \\
 & \mathcal{O}(\underline{\boldsymbol{x}},\underline{\boldsymbol{y}}^{<i-1}, \boldsymbol{y}^{i-1}),
    \label{objective}
\end{align}
where $|\boldsymbol{y}|$ denotes the number of tokens in the sentence $\boldsymbol{y}$, and where the base case $\mathcal{O}(\underline{\boldsymbol{x}},\underline{\boldsymbol{y}}^{<0}, \boldsymbol{y}^{0}) = 0$. Note that Eq.~\ref{objective} is a generalization of Eq.~\ref{iid_bayes} in $\log$ space---if we set $\lambda_1 = \lambda_3 = 0$ and $\lambda_2=1$ and take the $\log$ of Equation~\ref{iid_bayes} the two objectives are equivalent. 
The extra factors---the proposal probability and the length of the output---provide improvements (e.g.\ by calibrating the expected length of the output), and can be incorporated at no cost in the model; they are widely used in prior work \citep{moses:2007,yu:2017,yee-etal-2019-simple,DBLP:conf/wmt/NgYBOAE19}.
The elements on the beam after considering the $\ell$th sentence are reranked one final time by adding $\log p_{\textrm{LM}}(\langle \textsc{stop} \rangle \mid \underline{\boldsymbol{y}}^{\le \ell})$ to the final score; this accounts for the language model's assessment that the candidate document has been appropriately ended.\footnote{When sentences are modeled independently, this quantity is constant and can be ignored.}

\ignore{The coefficients on the extra factors---the proposal probability and the length of the output---are tuned to maximize an MT quality metric on a held-out validation set and provide improvements (e.g., by calibrating the expected length of the output).}
\ignore{The recursive form of Eq.~\ref{objective} emphasizes that the score of a prefix of a partial translation can be cached and only the contribution to the score from the next candidate sentence $\boldsymbol{y}^i$ needs to be evaluated.}

\section{Experiments}
\label{sec:experiments}

We evaluate our model on two translation tasks, the NIST Open MT Chinese--English task\footnote{\url{https://www.nist.gov/itl/iad/mig/open-machine-translation-evaluation}} and the WMT19 Chinese--English news translation task.\footnote{\url{http://www.statmt.org/wmt19/translation-task.html}} On both tasks, we use the standard parallel training data, and compare our model with a strong transformer baseline, as well as related models from prior work.

\subsection{Dataset Description}
The NIST training data is composed from LDC-distributed news articles and broadcast transcripts and consists of 1.5M sentence pairs. The document-level parallel corpus is a subset of the full training set, including 55K documents with 1.2M sentences. Following prior work, we use the MT06 dataset as validation set and MT03, MT04, MT05, and MT08 as test sets. There are 79 documents and 1649 sentences in the validation set and in total 509 documents and 5146 sentences in the test set. On average, documents in the test set has 10 sentences and 250 words and 330 words on the Chinese and English sides, respectively. We preprocess the dataset by doing punctuation normalization, tokenization, and lower casing. We use byte pair encoding \citep{SennrichHB16a} with 32K merges to segment words into sub-word units for both Chinese and English. The evaluation metric is case-insensitive BLEU calculated using {\tt multi-bleu.perl}, which is consistent with prior work on this task.

The training data for the WMT19 Chinese--English task includes the UN corpus, CWMT, and news commentary. The total number of sentence pairs is 18M after filtering the data by removing duplicate sentences and sentences longer than 250 words. The validation sets that we use in the experiment are newstest2017 and newstest2018, which contains 169 documents, 2001 sentences and 275 documents, 3981 sentences, respectively. The test set is newstest2019 containing 163 documents and 2000 sentences. On average, documents in the test set has 12 sentences and 360 words and 500 words on the Chinese and English sides, respectively. The dataset is preprocessed by segmenting Chinese sentences and normalizing punctuation, tokenizing, and true casing English sentences. As for NIST, we learn a byte pair encoding \citep{SennrichHB16a} with 32K merges to segment words into sub-word units for both Chinese and English. The evaluation metric is {\it sacreBLEU} \citep{post-2018-call}.

\subsection{Model Configuration}
For NIST, we use the transformer \citep{VaswaniSPUJGKP17} as the channel model and the document transformer \citep{ZhangLSZXZL18} as the proposal model. The hyperparameters for training the transformer are the same as {\it transformer base} \citep{VaswaniSPUJGKP17}, i.e.\ 512 hidden size, 2048 filter size, 8 attention heads, and 6 layers for both the encoder and decoder. We follow \citet{ZhangLSZXZL18}'s configuration to train the \textit{document transformer}: context length is set to 2 and all other hyperparameters are the same as {\it transformer base}. Both models are optimized using Adam \citep{KingmaB14} with approximately 24K BPE tokens per mini-batch.
For the language model, we train the transformer-XL \citep{DaiYYCLS19} on a combination of the English side of NIST training data as well as three sections of Gigaword: XIN, AFP, APW, resulting in a total of 7.3M documents and 115M sentences. We use an architecture with 24 layers, 16 attention heads, and embeddings of dimension 1024.
The input sequences to the language model are encoded into bytes using the byte-level encoder provided by GPT2~\citep{radford2019language}.

For WMT19, we use the transformer \ignore{\citep{VaswaniSPUJGKP17}} as both the channel and proposal model. The hyperparameters for training the transformer is the same as {\it transformer big} \citep{VaswaniSPUJGKP17}, i.e.~1024 hidden size, 4096 filter size, 16 attention heads, and 6 layers. The model is trained on 8 GPUs with batch size of 4096. The setup for the language model is the same as that of NIST except that the training data is the English side of the parallel training data and Gigaword.

For both tasks, the weights $\boldsymbol{\lambda}$ are selected using grid search, from $[0.8, 1., 1.5, 2., 2.2, 2.5, 3.]$ for the weights of channel model $\lambda_2$ and proposal model $\lambda_1$, and from $[0.2, 0.5, 0.8, 1.]$ for the length penalty $\lambda_3$. The size of the $n$-best list used in the reranker is set to $K=50$.\footnote{$K=50$ gives the best compromise between performance and inference time.} The beam size in the document decoding algorithm is $B=5$.

The running time for our decoding algorithm (Section \ref{sec:decoding}) highly depends on the language model's speed of calculating probabilities of partial documents. Using the transformer-XL language model with the aforementioned configuration, it takes approximately 90 seconds to decode a document on a Google Cloud TPU v3. We leave systematic exploration of inference algorithms for better solving the decoding problem to future work.

\subsection{Experimental Results}

\begin{table*}[t]\centering
    \small
	\begin{tabular}{@{}lccccccc@{}}
		\toprule
		Method & Model              & Proposal    & MT06  & MT03  & MT04  & MT05  & MT08  \\
		\midrule
		\citep{WangTWL17}      & RNNsearch      & -        & 37.76 & -     & -     & 36.89 & 27.57 \\
		\citep{kuang}       & Transformer + cache     & -       & 48.14 & 48.05 & 47.91 & 48.53 & 38.38 \\
		\citep{ZhangLSZXZL18}       & Doc-transformer  & -       & 49.69 & 50.21 & 49.73 & 49.46 & 39.69 \\
		\midrule
		\multirow{4}{*}{Baseline} & Sent-transformer  & -  & 47.72 & 47.21 & 49.08 & 46.86 & 40.18 \\
		       & Doc-transformer ($q$)  & -     & 49.79 & 49.29 & 50.17 & 48.99 & 41.70 \\
		 & Backtranslation ($q'$) & - & 50.77 & 51.80 & 51.61 & 51.81 & 42.47 \\
		 & Sent-reranker  & $q$  & 51.33  & 52.23 & 52.36 & 51.63 & 43.63 \\
		\midrule
		\multirow{2}{*}{This work}  & Doc-reranker      & $q$    &  51.99 & 52.77 & 52.84 & 51.84 & 44.17 \\  
		& Doc-reranker & $q'$ & \bfseries{53.63} & \bfseries{54.51} & \bfseries{54.23} & \bfseries{54.86} & \bfseries{45.17} \\
		 
		\bottomrule
	\end{tabular}
	\caption{Comparison with prior work on NIST Chinese--English translation task. The evaluation metric is tokenized case-insensitive BLEU. The first three rows are numbers reported in the papers of prior work. 
	The first two baselines are the results that we got by running the transformer \citep{VaswaniSPUJGKP17} and the document transformer \citep{ZhangLSZXZL18} on the NIST dataset. The sent-reranker is a variation of our model in which sentences in documents are assumed to be independent. The backtranslation baseline is obtained by training the document transformer using additional synthetic parallel documents generated by backtranslation.}
	\label{nist_prev_result}
\end{table*}

Table \ref{nist_prev_result} presents the best result from our model (doc-reranker) in comparison with prior work on the NIST Chinese--English translation task. The first three rows are numbers reported in prior work. \citet{WangTWL17} incorporate document context by introducing a hierarchical RNN to an LSTM sequence-to-sequence model. \citet{kuang} use a cache to store previously translated words across sentences, which they then use in sequence-to-sequence models. \citet{ZhangLSZXZL18} extend the transformer model with an extra context encoder to capture information from previous source sentences. Apart from prior work, we also compare our doc-reranker with four baselines: the transformer \citep{VaswaniSPUJGKP17}, document transformer \citep{ZhangLSZXZL18}, the sentence-level reranker (sent-reranker), and the document transformer with backtranslation.

In the sent-reranker, we assume sentences in the document are independent (formulation $
    \hat{\underline{\boldsymbol{y}}} = \arg \max_{\underline{\boldsymbol{y}}} \prod_{i=1}^{|\underline{\boldsymbol{x}}|} p(\boldsymbol{x}^i \mid \boldsymbol{y}^i) \times p(\boldsymbol{y}^i)
$), and therefore we train a sentence-level language model and rerank each sentence independently. This sent-reranker setup is close to the work from \citet{yee-etal-2019-simple} and \citet{DBLP:conf/wmt/NgYBOAE19} with the difference that rather than using a language model trained on documents we use a language model trained on sentences, which is more statistically consistent.
\ignore{The numbers corresponding to these baseline models in Table \ref{nist_prev_result} are obtained by running the models on our processed NIST data.}Table \ref{nist_prev_result} shows that our reranker outperforms previous models as well as strong transformer baselines by a significant margin---approximately 2.5 BLEU on all test sets---achieving new state of the art. Although the gap between the doc-reranker and sent-reranker is smaller, as we will show in \S\ref{sec:human_eval} and \S\ref{sec:analysis} that translations generated by doc-reranker are preferred by humans and are more consistent across documents, in line with concerns about the reliability of using BLEU at assessing cross-sentential consistency~\citep{voita-etal-2019-good}.

To compare the effectiveness of leveraging monolingual data between backtranslation \citep{EdunovOAG18,DBLP:conf/acl/SennrichHB16} and our model, we train the document transformer \citep{ZhangLSZXZL18} using additional synthetic parallel documents generated by backtranslation $(q')$. For fair comparison we use the same monolingual data for both models. As shown in Table \ref{nist_prev_result} that while both techniques improve translation backtranslation is less effective than our model. Since we have a new model $q'$, we can use it as a proposal model for our doc-reranker---effectively using the monolingual data twice. We find that this improves results even further, indicating that the effect of both approaches is additive.

\begin{table*}\centering
    \small
	\begin{tabular}{@{}lccccc@{}}
		\toprule
		 Proposal model & Language model  & Sent-reranker & Doc-reranker \\
		\midrule
		\multirow{3}{*}{Sent-transformer} &  LSTM: NIST & 49.92 & 50.24 \\
		 &  transformer-XL: NIST & 50.29 & 50.56 \\
	     & transformer-XL: NIST + Gigaword & 50.19 & 50.93 \\
	    \midrule
	    \multirow{3}{*}{Doc-transformer} &  LSTM: NIST & 50.75 & 51.20 \\
	     & transformer-XL: NIST & 51.27 & 51.68 \\
		 & transformer-XL: NIST + Gigaword & 51.33 & 51.99 \\
		\bottomrule
	\end{tabular}
	\caption{BLEU scores on NIST dev set MT06 from rerankers which are incorporated with various language models. In the language model column X: Y means the language model X is trained on dataset Y. A bigger language model improves the doc-reranker but does not help the sent-reranker.}
	\label{nist_result}
\end{table*}

\begin{table}[h]
    \centering
    \small
	\begin{tabular}{@{}lllccccc@{}}
		\toprule
		 Architecture & Data & PPL \\
		\midrule
		 transformer-XL & NIST sent & 83.3 \\ 
		 transformer-XL & NIST + GW sent & 96.5 \\ 
		 \midrule
		 LSTM & NIST doc & 71.6\\
		 transformer-XL & NIST doc & 43.8 \\ 
		 transformer-XL & NIST + GW doc & 43.4 \\ 
		\bottomrule
	\end{tabular}
	\caption{Perplexity per word of  language models on NIST dev set. GW refers to Gigaword.}
	\label{lm_ppl}
\end{table}

\begin{table}[h]
    \centering
    \small
    \begin{tabular}{@{}llc@{}}
	\toprule
	Reranker        & Models                    & MT06  \\
	\midrule 
	-               & Doc-transformer           & 49.79 \\
	\midrule
	\multirow{4}{*}{Doc-reranker} & Proposal + LM &  49.79   \\
		          & Channel + LM              &   51.93    \\
		          & Proposal + Channel        &   50.40    \\
		          & Proposal + Channel + LM  & \bfseries{51.99} \\
		
	\bottomrule
	\end{tabular}
    \caption{Effect of different components.\ignore{in doc-reranker.}}
    \label{nist_ablation}
\end{table}

\begin{figure}[h]
	\centering
	\includegraphics[scale=0.40]{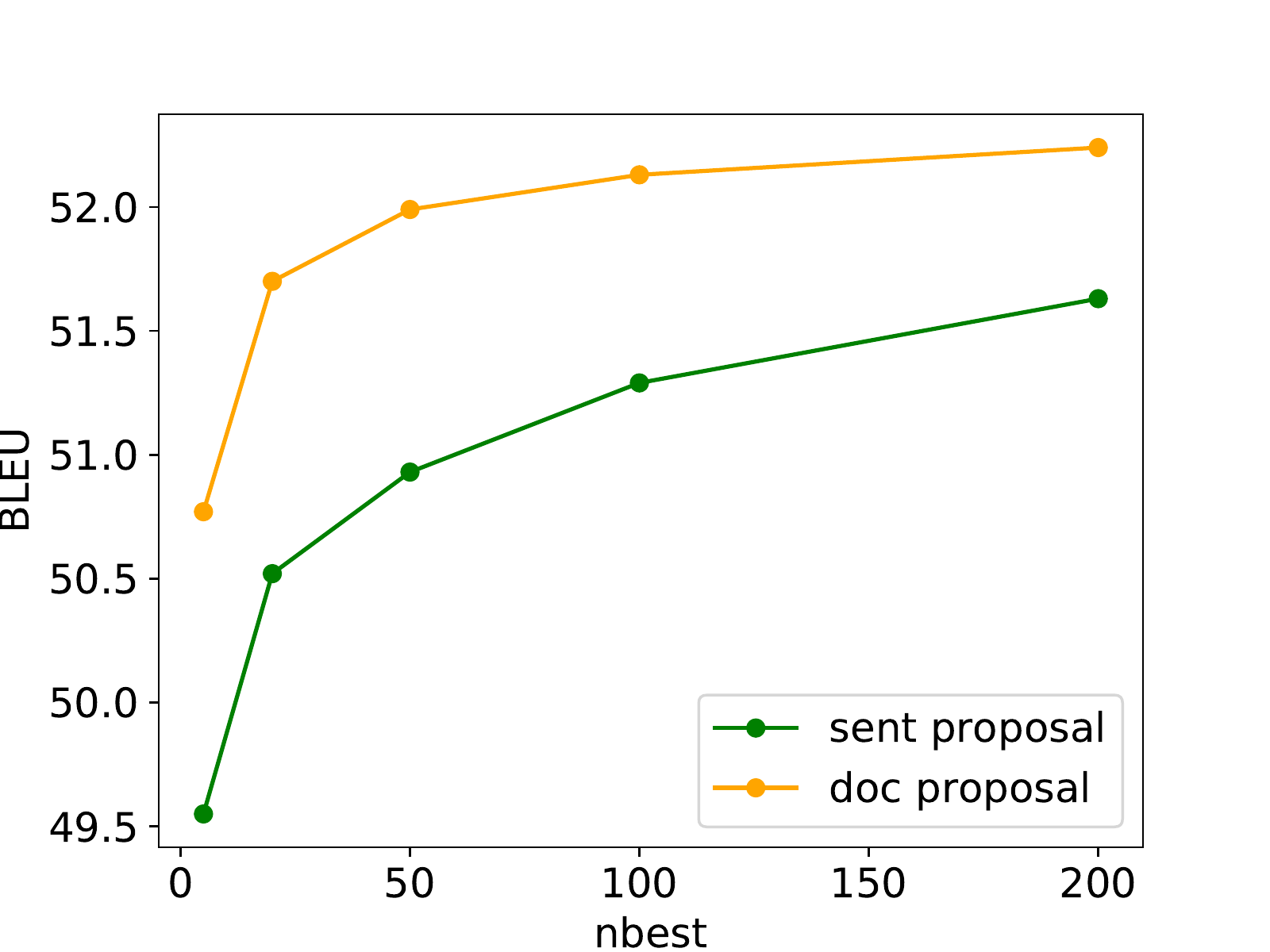}
    \caption{Effect of {\it n}-best list.}
    \label{bleu_curve}
\end{figure}

To understand the rerankers better, we investigate the effect of different proposal models, different language models, and various numbers of candidates in the $n$-best list. Table \ref{nist_result} and Figure \ref{bleu_curve} show that better proposal models and bigger $n$-best lists lead to consistently better reranking results. This is an appealing behaviour showing that our reranker is able to pick better translations from higher quality and more diverse candidate pools generated by better proposal models and bigger $n$-best lists. To compare the effect of language models, we train an LSTM language model \citep{DBLP:conf/iclr/MerityKS18,DBLP:journals/corr/abs-1803-08240} and a transformer-XL language model on the English side of NIST parallel training data in addition to the transformer-XL trained on NIST and Gigaword. Table \ref{lm_ppl} lists the perplexity per word on the NIST validation set for different language models. Given the same training data, the transformer-XL performs significantly better than the LSTM-based language model, which in turn results in a higher BLEU score from the doc-reranker. By adding more training data, the transformer-XL language model achieves even lower perplexity and that gives a further boost to the performance of the doc-reranker. Notably, when the strong transformer-XL language model is incorporated into the doc-reranker, the best weight ratio of the channel and language model is $1:1$, indicating that the doc-reranker depends heavily on the language model. By contrast, if a weak language model is incorporated, the best ratio is approximately $2:1$. A further observation is that although a larger-scale language model improves the doc-reranker, it does not help the sent-reranker.

We perform an ablation study to explore what each component of the doc-reranker contributes to the overall performance. Table~\ref{nist_ablation} shows BLEU scores on the NIST validation set for the optimal interpolation of various component models. No gains are observed if the language model is combined with the proposal model (a probabilistically unsound combination, although one that often worked in pre-neural approaches to statistical translation). We find that as we increase the weight of the language model, the results become worse. The interpolation of the proposal model and channel model slightly outperforms the proposal model baseline but considerably underperforms the interpolation of the proposal model, channel model, and the language model. This difference indicates the key roles that the language model plays in the doc-reranker. When the channel model is combined with the language model the performance of the doc-reranker is comparable to that with all three components included.\ignore{It is expected that the proposal model does not add new information to the objective if Bayes' rule yields a better estimate of the translation probability than its direct estimation. Thus, since we believe the channel model and language model to be well estimated, we expect that the extra proposal model to add little value.}  
We conclude from the ablation study that both the channel and language models are indispensable for the doc-reranker, indicating that Bayes' rule provides reliable estimates of translation probabilities.

\begin{table*}[t]\centering
    \small
	\begin{tabular}{@{}lccccccc@{}}
		\toprule
		Method                & Model   & Unpaired Data & LM PPL & Test17 & Test18 & Test19
		\\
		\midrule
		Baseline              & transformer big & - & - & 23.9   & 23.9         & 24.5 \\
		\midrule
		\multirow{2}{*}{This work}  & \multirow{2}{*}{Doc-reranker}  & WMT & 106.3 & 24.9 & 26.0 & 27.1 \\
		&  & Gigaword + WMT & 63.8 & \bfseries{25.5} & \bfseries{26.3} & \bfseries{27.1} \\
		\bottomrule
	\end{tabular}
	\caption{SacreBLEU of different models on WMT19 validation and test sets and perplexity per word of the language models on the English side of WMT19 validation set.}
	\label{wmt_result}
\end{table*}

Table \ref{wmt_result} presents the results of our model together with baselines on the WMT19 Chinese--English translation task. We find that the results follow the same pattern as those on NIST: a better language model leads to better translation results and overall the reranker outperforms the transformer-big by approximately 2.5 BLEU. The two best systems submitted to the WMT19 Chinese--English translation task are Microsoft Research Asia's system \citep{xia2019microsoft} and Baidu's system \citep{sun2019baidu}, both of which employ multiple techniques to improve upon the transformer big model. 
Here we mainly compare our results with those from \citet{xia2019microsoft} because we use the same evaluation metric {\it SacreBLEU} \citep{post-2018-call} and the same validation and test sets.
Using extra parallel training data and the techniques of masked sequence-to-sequence pretraining \citep{DBLP:conf/icml/SongTQLL19}, sequence-level knowledge distillation \citep{KimR16}, and backtranslation \citep{EdunovOAG18}, the best model from \citet{xia2019microsoft} achieves 30.8, 30.9, and 39.3 on newstest2017, newstest2018, and newstest2019, respectively. Although our best results are lower than this, it is notable that our model achieves comparable results to their model trained on 56M sentences of parallel data, over two times more training data than we use. However, our method is orthogonal to these works and can be combined with other techniques to make further improvement.

\section{Analysis}
In this section, we present the quantitative and qualitative analysis of our models. The analysis is performed on the experiments of the NIST dataset.
\subsection{Quantitative Analysis}
 
\begin{figure}
	\centering
	\includegraphics[scale=0.40]{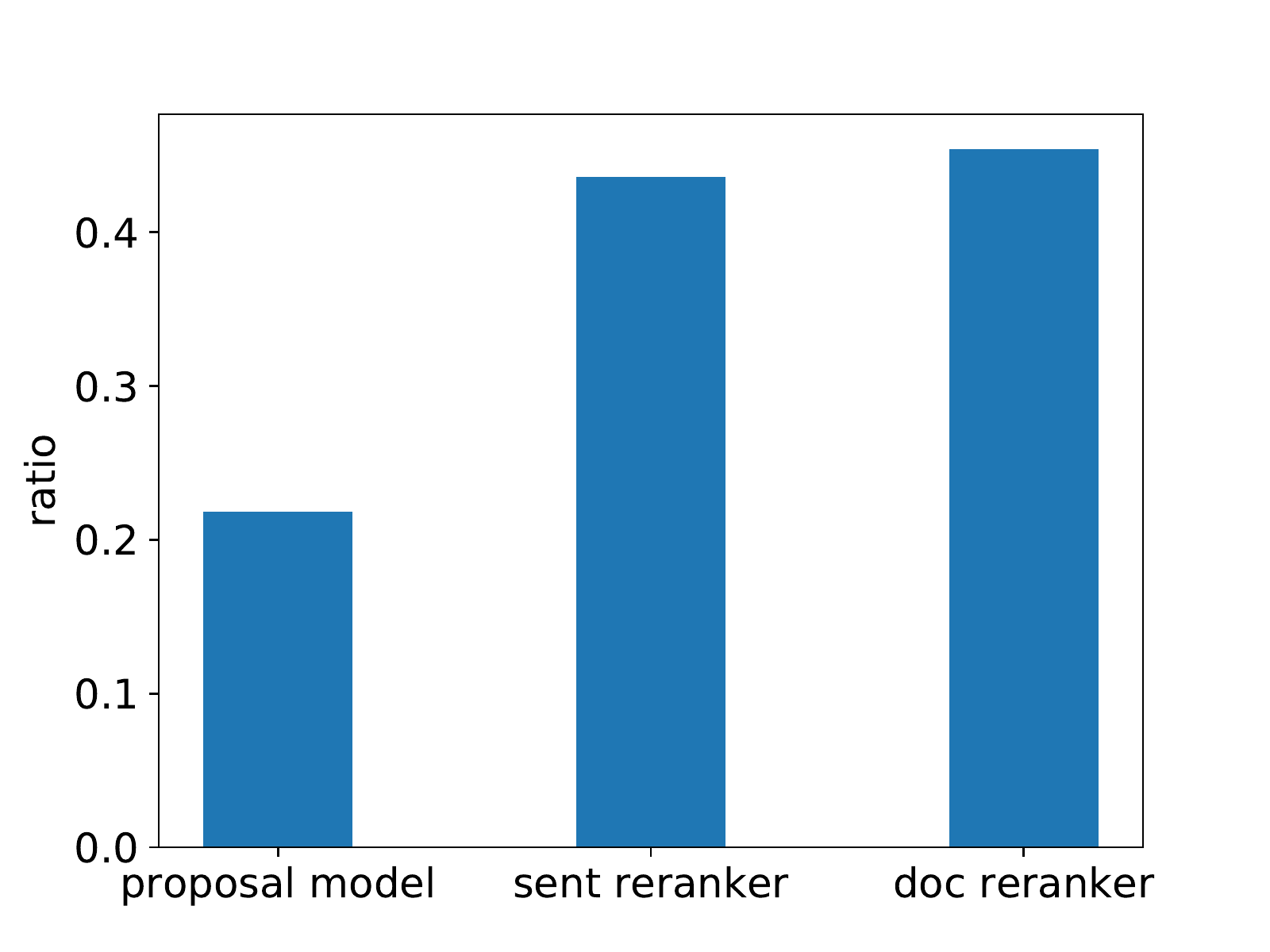}
    \caption{Ratio of different models picking true targets.}
    \label{ratio_histogram}
\end{figure}

We do oracle experiments in order to assess our models' ability to select good translation candidates. We create our candidate pool by mixing the proposals generated from the transformer model \citep{VaswaniSPUJGKP17} and the four references. We subsequently calculate how many cases over the entire validation dataset in which different models (the proposal model, sent-reranker, and doc-reranker) assign the highest model scores to the reference translations. As shown in Figure \ref{ratio_histogram} while the proposal model selects one of the references as the best candidate for 22\% of the sentences in the validation dataset, both rerankers double the ratio and the doc-reranker achieves 2\% higher accuracy than the sent-reranker. This observation provides a further evidence that if we improve the quality of the candidate pool our model will generate better translations. 

\begin{table}\centering
    \small
	\begin{tabular}{@{}lcccc@{}}
		\toprule
		 Proposal & \#Experts & pBLEU & BLEU \\
		\midrule
	     human  & 4 & 21.40 & - \\ 
		\midrule
		 \multirow{3}{*}{Doc-transformer} & 1 & 70.41 & 53.63 \\
		  & 2 & 59.09 & 54.70 \\ 
		  & 4 & \bfseries{53.54} & \bfseries{55.21} \\ 
		\bottomrule
	\end{tabular}
	\caption{Pairwise-BLEU (pBLEU) \citep{ShenOAR19} for candidate translations generated from different number of experts. BLEU from the doc-reranker taking different sets of candidate translations. We obtain different experts by training the document transformer \citep{ZhangLSZXZL18} with backtranslation with different random initialization. The size of candidate pool is 50. The experts for the human proposal baseline are the reference translations. }
	\label{diversity}
\end{table}

We also assess the diversity of the candidate pool and investigate the effect of their diversity on our model's performance. Table \ref{diversity} lists pairwise-BLEU\footnote{Pairwise-BLEU \citep{ShenOAR19} is a metric of measuring the similarity of candidate translations. The lower the pairwise-BLEU is, the more diverse the candidate translations are. We refer the readers to \citet{ShenOAR19} for the definition of the metric.} scores \citep{ShenOAR19} of different candidate pools (of size 50) and their corresponding BLEU scores from the doc-reranker. We use the document transformer \citep{ZhangLSZXZL18} trained with additional backtranslated synthetic documents as the proposal models ($q'$ in Table \ref{nist_prev_result}) in the doc-reranker. Table \ref{diversity} shows that the candidates generated from our proposal model (by taking 50 best sentences from the beam search) are much less diverse than human translations. We conjecture that the lack of diversity in the candidate pool may harm the performance of our model.

To increase the diversity of candidate translations, we create candidate pools by composing translations generated from different ``experts'', which are simply document transformer models trained from different random initializations. As illustrated in Table \ref{diversity}, we find that a candidate pool from more experts results in more diverse translations (quantified by pairwise BLEU) and better reranking results (quantified by BLEU).

\subsection{Qualitative Analysis}
\label{sec:analysis}
To investigate how the rerankers improve translation quality, we analyze the output from different models: the document transformer \citep{ZhangLSZXZL18} (our proposal model), the sent-reranker, and the doc-reranker. We observe that in general the  doc-reranker improves adequacy of translations\ignore{(in line with the findings by \citet{yu:2017})} and can generate more fluent and natural sentences compared to the document transformer. More importantly, our doc-reranker shows its superiority over the others in terms of exploiting context, improving consistency of tense, number, and lexical choice across entire articles.  Tables \ref{tb:proposal_vs_reranker} and \ref{tb:sent_vs_doc} in Appendix \ref{sec:app} present example output from the aforementioned systems. In Example 1, the pronoun {\it he} is omitted in the Chinese sentence. While the document transformer misses this pronoun resulting in a translation of completely different meaning, the doc-reranker is able to recover it. Likewise, in Example 6 {\it them} is dropped in the source sentence and this pronoun can only be inferred from previous context. Although both rerankers recover some pronoun, only the doc-reranker gets it right, by relying on cross-sentential context. Example 2 is a good example showing that the doc-reranker is better at generating adequate translations than the proposal model: the document transformer ignores the phrase {\it with these people}, but the doc-reranker covers it.

Chinese does not mark nouns for number, and it therefore has to be inferred from context to generate accurate English translations.
It is not possible for a sentence-level MT system to capture this information if the relevant context is not from the current sentence. In Example 3 and 5 the plural {\it problems} and {\it identities} can only be inferred from previous sentences (the immediate previous sentence in Example 3 and the sentence 4-5 sentences away from the current one in Example 5). While neither the document transformer nor the sent-reranker makes the right predictions in both examples, the doc-reranker translates correctly, indicating its strength in capturing extra-sentential information. In addition to making inference across sentences, the doc-reranker is also capable of maintaining consistency of tense and lexical choice, as demonstrated in Examples 4, 7, and 9. Furthermore, it improves the consistency of writing style.
To illustrate, in Example 8, the context is that of a list of bullet points starting with {\it continue}.
The doc-reranker follows in this style by starting the translation with the verb {\it continue}. However, the sent-reranker starts the sentence with {\it we should continue}. Although both translations are reasonable, the former one is more natural within the document since it preserves stylistic consistency.

\section{Related Work}
Our work is closely related to three lines of research: context-aware neural machine translation, large-scale language models for language understanding, and semi-supervised machine translation. Recent studies \citep[\textit{inter alia}]{TiedemannS17,BawdenSBH18} have shown that exploiting document-level context improves translation performance, and in particular improves lexical consistency and coherence of the translated text. Existing work in the area of context-aware NMT typically adapts the MT system to take additional context as input, either a few previous sentences \citep{JeanLFC17,WangTWL17,TuLSZ18,TitovSSV18,ZhangLSZXZL18,WerlenRPH18} or the full document \citep{HaffariM18,MarufMH19}. 
These methods varies in the way of encoding the additional context and the way of integrating the context with the existing sequence-to-sequence models. For example, \citet{WerlenRPH18} encode the context with a separate transformer encoder \citep{VaswaniSPUJGKP17} and use a hierarchical attention model to integrate the context into the rest of transformer model. \citet{ZhangLSZXZL18} introduce an extra self-attention layer in the encoder to attend over the the context.

Strategies for exploiting monolingual document-level data have been explored in two recent works \citep{voita-etal-2019-context,Junczys-Dowmunt19}. Both use backtranslation \citep{EdunovOAG18,DBLP:conf/acl/SennrichHB16} to create synthetic parallel documents as additional training data. In contrast, we train a large-scale language model and use it to refine the consistency between sentences under a noisy channel framework. Advantages of our model over backtranslation are that 1) the language model is portable across domain and language pairs; 2) our model involves straightforward training procedures. Specifically, for backtranslation to succeed, monolingual data that will be back translated must be carefully selected; the ratio of backtranslated data and original data must be balanced carefully. While techniques for doing this are fairly well established for single sentence models, no such established techniques exist for documents.  

More generally, strategies for using monolingual data in nueral MT systems is an active research area \citep[\textit{inter alia}]{DBLP:journals/corr/GulcehreFXCBLBS15,DBLP:conf/acl/ChengXHHWSL16}. 
Backtranslation \citep{EdunovOAG18,DBLP:conf/acl/SennrichHB16}, originally invented for semi-supervised MT, has been employed as a standard approach for unsupervised MT \citep{Lample/emnlp:18,Lample/iclr:18,DBLP:conf/acl/ArtetxeLA19,DBLP:conf/iclr/ArtetxeLAC18}. Noisy channel decompositions have been a standard approach in statistical machine translation \citep{DBLP:journals/coling/BrownPPM94,moses:2007} and recently have been applied to neural models \citep{yu:2017,yee-etal-2019-simple,DBLP:conf/wmt/NgYBOAE19}. 
Unlike prior work, we adopt noisy channel models for document-level MT. While the model from \citet{yu:2017} could be used on documents by concatenating their sentences to form a single long sequence, this would not let us use the conditional sentence independence assumptions that gives our model the flexibility to use just parallel sentences. Secondarily, their inference algorithm is specialized to their channel model, and it has a quadratic complexity, which would be prohibitive for sequence longer than a single sentence; in practice our inference technique is much faster.

Large-scale pretrained language models have achieved success in improving systems in language understanding, leading to state-of-the-art results on a wide range of tasks \citep{Peters:2018,DBLP:conf/naacl/DevlinCLT19,radford2018improving,DBLP:conf/nips/McCannBXS17,DBLP:journals/corr/abs-1906-08237,DBLP:conf/naacl/ChronopoulouBP19,DBLP:journals/corr/abs-1901-07291}.\ignore{The paradigm is that transformer \citep{VaswaniSPUJGKP17} or RNN models are trained on a large corpus in an unsupervised fashion and then fine-tuned on supervised data in downstream transfer tasks. A series of follow-up work has optimized this approach \citep{DBLP:journals/corr/abs-1906-08237,DBLP:conf/naacl/ChronopoulouBP19} and extended it to cross-lingual understanding \citep{DBLP:journals/corr/abs-1901-07291}.} Language generation is another area where pretrained language models have been applied, with existing work focusing on fine-tuning for repurposing an unconditional language model \citep{DBLP:journals/corr/abs-1902-09243,DBLP:conf/naacl/EdunovBA19,DBLP:conf/icml/SongTQLL19,DBLP:journals/corr/abs-1905-03197,ziegler:2019,Oliveira2019RepurposingDL}. In contrast to our work which uses probabilities from language models, that work uses model internal representations. 

\section{Conclusion}
We have presented a noisy channel reranker and empirically validated it on Chinese--English document-level translation tasks. The noisy channel formulation requires only parallel sentences (rather than documents) but we can use abundant monolingual documents to train the language model component. Experiments show that our proposed model considerably improves translation quality---it achieves approximately 2.5 BLEU higher than transformer baselines. Subjective evaluation further confirms that the language model helps enforce consistency of tense, number, and lexical choice across documents.

\appendix
\section{Appendix}
\label{sec:app}
\subsection{Human Evaluation}
\label{sec:human_eval}

We selected 50 translation triplets (reference translation, translation from the doc-reranker, translation from the sent-reranker) sampled from the validation and test sets of NIST for evaluation by 4 native English speakers. The samples are selected by taking the triplets where the output from the sent-reranker and the doc-reranker have a translation edit rate~\citep{snover:2006} above 17.5\%. 

Each of these documents was presented with a reference translation and with two translations labelled A and B, one generated by the doc-reranker and one generated by the sent-reranker. They were tasked with indicating which of these two they found better overall, considering fluency, idiomaticness and correctness (relatively to the reference).

Each of the human evaluators preferred a majority of doc-reranker translations. When aggregated for each document by majority vote, the doc-reranker translations were considered better in 25 documents, worse for 13, and tied for 12. A statistically significant preference at $p < 0.05$ according to an exact one-tailed Binomial test ($n = 38$).

\subsection{Comparison of Output from Different Systems}
To investigate how the rerankers improve translation quality, we manually inspect the output from three different systems: the document transformer \citep{ZhangLSZXZL18}, the sent-reranker, and the doc-reranker. 
Tables \ref{tb:proposal_vs_reranker} and \ref{tb:sent_vs_doc} present the comparison between the output from the document transformer \citep{ZhangLSZXZL18} and sent-reranker and between the output from sent-reranker and doc-reranker, respectively. In general, we find that the doc-reranker outperforms other systems in terms of maintaining consistency of tense, number, and lexical choices across documents. For detailed analysis, we refer readers to \S\ref{sec:analysis}.

\begin{table*}
    \footnotesize
    \centering
	\begin{tabular}{@{}llp{14.2cm}@{}}
		\toprule
		1 & \bfseries{src:} & \begin{CJK*}{UTF8}{gbsn}
			霍夫曼在接受美国哥伦比亚广播公司新闻杂志「六十分钟」访问时轻叹,那段时期为了得到毒品和酒,真是不择手段。
		\end{CJK*}  \\
		& \bfseries{ref:} & in an interview on us cbs news magazine 60 minutes, hoffman softly sighed that in such period \underline{he} would truly do anything to get drugs and alcohol. \\
		& \bfseries{out1:} & in an interview with the cbs news magazine ``60 minutes", hoffmann sighed that \underline{those days were} really unscrupulous in getting drugs and alcohol. \\
		& \bfseries{out2:} & in an interview with the cbs news magazine ``60 minutes", hoffmann sighed that at that time in order to obtain drugs and alcohol, \underline{he} was really unscrupulous. \\
		\midrule
		2 
		& \bfseries{ref:} & in the meantime, more than 10 chinese personnel \underline{working in the same place} \underline{with these people} have been called back to karachi. at present they are emotionally stabilized. \\
	    &	\bfseries{out1:} & at the same time, more than ten chinese personnel \underline{working at the same site} have also withdrawn to karachi. their sentiments are now stable. \\
		& \bfseries{out2:} & at the same time, more than ten chinese personnel \underline{working with these people on the same site} have also withdrawn to karachi. at present, their sentiments are stable. \\
		\midrule
		3 & \bfseries{src:} & \begin{CJK*}{UTF8}{gbsn}
		    基本的问题是什么呢?
		\end{CJK*}\\
	    & \bfseries{cxt:} & \dots however, legislator yeung, i wish to tell you what i am doing today is to ensure every matter can proceed smoothly after the political review starts. therefore, we have to \underline{solve some basic problems} first and this is a different thing all together. \\
		& \bfseries{ref:} & what \underline{are the basic problems}? \\
		& \bfseries{out1:} & what \underline{is the basic problem}? \\
		& \bfseries{out2:} & what \underline{are the basic questions}?  \\
		\midrule
		4 
	    & \bfseries{cxt:} & sword of \underline{justice}: prospects for 2006 \\
		& \bfseries{ref:} & author: sword of \underline{justice} \\
		& \bfseries{out1:} & author: the sword of \underline{righteousness} \\
		& \bfseries{out2:} & author: the sword of \underline{justice}  \\
		\bottomrule
	\end{tabular}
	\caption {Example outputs from the document transformer (out1) and our doc-reranker (out2).}
	\label{tb:proposal_vs_reranker}
\end{table*}

\begin{table*}
    \footnotesize
    \centering
	\begin{tabular}{@{}llp{14.2cm}@{}}
		\toprule
		5 & \bfseries{src:} & \begin{CJK*}{UTF8}{gbsn}
		同时我们在国内用最短的时间，核实清楚了死亡人员的身份。
		 \end{CJK*}  \\
		& \bfseries{cxt:} & \dots the criminal used a submachine gun to fire a barrage of shots, and \underline{three engineers} died unfortunately. \dots \\
		& \bfseries{ref:} &  at the same time, we in china verified the \underline{identities} of the dead within the shortest possible time. \\
		& \bfseries{out1:} & at the same time, we spent the shortest time in china to verify the \underline{identity} of the deceased. \\
		& \bfseries{out2:} & at the same time, we spent the shortest time in china to verify the \underline{identities} of the deceased.\\
		\midrule
		6 & \bfseries{src:} & \begin{CJK*}{UTF8}{gbsn}
			现在又要平安的送到家里。
		\end{CJK*} \\
		& \bfseries{cxt:} & \dots when the plane carrying the  \underline{three survivors} and 11 other personnel arrived in Hefei, people waiting at the airport heaved a long sigh of relief. \dots after the incident occurred, it made proper arrangements for them.
        \\
		& \bfseries{ref:} &  now \underline{they} will also be escorted home safely. \\
		& \bfseries{out1:} & now they have to send \underline{it} home safely. \\
		& \bfseries{out2:} & now they want to send \underline{them} safely to their homes. \\
		\midrule
		7 &
		\bfseries{cxt:} & \dots a traffic accident \underline{occurred} at the 58 kilometer point of the beijing-harbin highway, with a spill from an oil tanker leading to the closure of a section of the highway. \dots it was \underline{learned} that the oil tanker contained waste oil from charcoal production. \dots\\
		& \bfseries{ref:} & the section of the highway from harbin to shuangcheng \underline{was} closed, with many vehicles detoured. \\
		& \bfseries{out1:} & part of the roads heading towards shuangcheng in harbin \underline{are} closed, and many vehicles are bypassing.\\
		& \bfseries{out2:} & part of the road from harbin to shuangcheng was closed , and many vehicles \underline{were} bypassing.\\
		\midrule
		8 & 
		\bfseries{cxt:} &  \dots with regard to coalmine safety this year, saws will effectively carry out the following three tasks: --\underline{continue} to effectively tackle the tough issue of controlling methane. \dots \\
		& \bfseries{ref:} & -- \underline{continue} to effectively tackle the tough issue of restructuring and shutting down. \\
		& \bfseries{out1:} & -- \underline{we should continue} to make a success of the rectification and closure battle. \\
		& \bfseries{out2:} &  -- \underline{continue} to fight the battle of rectification and closure.\\
		\midrule
		9 &
		\bfseries{cxt:} & \dots first, such abuse of  \underline{``quota"} restricts the thorough implementation of world trade organization's free trade principle. on one hand, u.s. is talking in high-sounding tone about ``free trade". on the other hand, it re-establishes trade barriers and stabs your back at will with \underline{``quotas"}. does it appear too arbitrary and unfair?  \\
		& \bfseries{ref:} & second, \underline{``quota"} limits the nice growth trend in sino-america trade relation. \\
		& \bfseries{out1:} & second, the \underline{``restriction"} restricts the good development momentum of sino-us economic and trade relations. \\
		& \bfseries{out2:} & second, the \underline{``quota"} restricts the good development momentum of sino-us economic and trade relations. \\
		\bottomrule
	\end{tabular}
	\caption {Example outputs from the sent-reranker (out1) and the doc-reranker (out2). cxt refers to context.}
	\label{tb:sent_vs_doc}
\end{table*}

\section*{Acknowledgement}
We would like to thank G\'{a}bor Melis for helpful comments on an
earlier draft of this paper and the language team at DeepMind for valuable discussions.

\bibliography{tacl2018}
\bibliographystyle{acl_natbib}

\end{document}